\documentclass[review]{elsarticle}

\usepackage{nopageno}
\usepackage{hyperref}
\usepackage{makecell}
\usepackage{array}
\journal{ArXiv.org}

\begin{document}

\begin{frontmatter}

\title{Automatic Model Building in GEFCom 2017 Qualifying Match}

\author[mymainaddress, mysecondaryaddress]{J{\'a}n Dolinsk{\'y}\corref{mycorrespondingauthor}}
\ead{jan.dolinsky@tangent.works}

\author[mymainaddress]{M{\'a}ria Starovsk{\'a}}
\cortext[mycorrespondingauthor]{Corresponding author}
\ead{maria.starovska@tangent.works}

\author[mymainaddress]{Robert T{\'o}th}
\ead{robert.toth@tangent.works}

\address[mymainaddress]{Tangent Works, Na Slav{\'i}ne 1, Bratislava, Slovakia}
\address[mysecondaryaddress]{Tangent Works, Oplombeekstraat 6, 1755 Gooik, Belgium}

\begin{abstract}
The Tangent Works team participated in GEFCom 2017 to test its automatic model building
strategy for time series known as Tangent Information Modeller (TIM).
Model building using TIM combined with historical temperature shuffling
resulted in winning the competition. This strategy involved one remaining degree of freedom,
a decision on using a trend variable.
This paper describes our modelling efforts in the competition, and
furthermore outlines a fully automated scenario where the decision on using
the trend variable is handled by TIM. The results show that such a setup would
also win the competition.
%Achieving the 1st place involved still one degree of freedom which was a decision on using a trend variable.
%The obtained 1st place shows that full automation is not only possible but is capable of delivering high quality results in accuracy too.
% This paper describes our modelling efforts in GEFCom 2017 in more detail.
% outperforming manually build models. This paper describes the modelling with TIM
%in GEFCom 2017 in more detail.
\end{abstract}


\end{frontmatter}

%\linenumbers

\section{Qualifying Match Description}

\begin{figure}[ht]
\label{Fig_LoadExample}
\caption{An example of forecast quantiles for the total consumption of the all ISO New England zones in January 2017.
Historical actuals are plotted in black. Quantile forecasts are plotted in grey.}
\centering
\includegraphics[width=11cm, height=4cm]{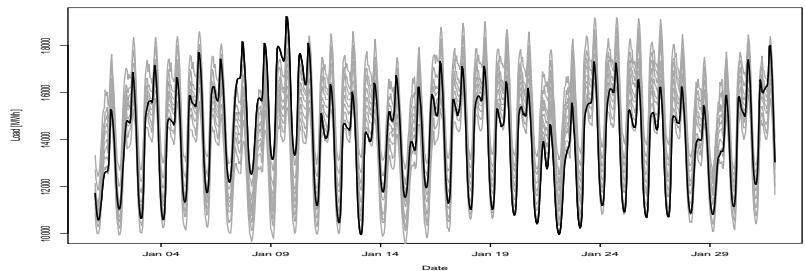}
\end{figure}

\paragraph{Data} The qualifying match consisted of making quantile predictions for 10 different time series of electricity load
sampled hourly (the 8 ISO New England zones, Massachusetts (sum of three zones under Massachusetts)
and the total (sum of the first 8 zones); see Table~\ref{Tab_ComparisonWithVanilla} for a ful list).
Fig.~\ref{Fig_LoadExample} shows an example of a load signal and coresponding quantile forecasts for January 2017.
Explanatory variable candidates were restricted to four - dry bulb temperature, dew point temperature, public holidays in the United States
and time information (time stamps for calculating month of year, day of week etc.).
Each forecast consisted of nine deciles (from first to ninth) and was submitted 6 times:
\begin{itemize}
\item Round 1 due date: Dec 15, 2016; forecast period: Jan 1-31, 2017.
\item Round 2 due date: Dec 31, 2016; forecast period: Feb 1-28, 2017.
\item Round 3 due date: Jan 15, 2017; forecast period: Feb 1-28, 2017.
\item Round 4 due date: Jan 31, 2017; forecast period: Mar 1-31, 2017.
\item Round 5 due date: Feb 14, 2017; forecast period: Mar 1-31, 2017.
\item Round 6 due date: Feb 28, 2017; forecast period: Apr 1-30, 2017.
\end{itemize}
Historical values were available for the year 2003 up to 2016.
There were no temperatures specified for the total Massachusetts load,
so we used 6 temperatures (2 temperatures from each of the 3 Massachusetts zones).

\paragraph{Preprocessing} The data quality was good and there was no preprocessing needed on the loads nor on the temperatures.
A slight change was done only to standardize the hours at the beginning and end of daylight saving time (DST).
The switch to DST in March required the averaging of the hour ending at 1AM and at 3AM so that a value of load ending at 2AM is created.
In November, the return to Standard Time is handled by halving the double counted 2AM value.
These modifications were not needed for the data since 2016.
This way we obtained 9 datasets with 4 variables (a vector of date strings, the 2 temperatures, and a vector
with ones for holidays and zeros otherwise) and 1 dataset (Massachusetts) with 8 variables.
Each dataset had 24 samples for each day in the years 2003 to 2016.

% rename to "Modelling Strategy" and promote to a section and merging with "Shuffling Temperatures" ?
\paragraph{Strategy} The task was defined as a probabilistic forecast. We, however,
decided not to use any of the modelling techniques created to handle these types of
forecast such as quantile regression \cite{Gaillard2016} or quantile regression forests.
Instead, we followed a strategy proposed in \cite{Tao2016}: \begin{enumerate} \item For
each dataset, create an underlying model that would score a high accuracy provided that
the exact values for temperatures would be known for each hour. \item Shuffle historical
temperatures to create 'new' artificial possible weather scenarios for forecast months and
plug them into these underlying models to obtain different load forecasts. Create
quantiles from these forecasts. \end{enumerate}

\section{Shuffling Temperatures} \noindent Luckily there were enough years of history
available for both dry bulb and dew point temperature - from 2004 to 2016, 13 years in
total (temperature from 2003 was not complete). This means we did not have to use
discontinuous shuffling to obtain a sufficient amount of new temperatures. What we did is
to simply shift the existing 13 year temperature profiles by 3 days ahead and 3 days
behind. This way we could generate 91 (7 times 13) 'new' temperatures, which is enough to
make reasonable quantiles from load after plugging the temperature into the underlying
model. These were simply constructed using the R function quantile, which calculates them
as $$Q[i](p) = (1 - \gamma) x[j] + \gamma x[j+1],$$ where all parameters are default taken.

\section{Model Building}
\noindent For each dataset only 2 models were created. These
models did not interfered with models created for other time series. The procedure of
creating the two models was identical in all 10 cases, meaning it suffices to describe only
one of them.

\paragraph{Training Data Ranges} Models were trained only on the last three years
(1.1.2014 to 30.11.2016). We retrained these models every time new data were available by
simply adding them to the dataset. In rounds 1, 2 and 3 data were available until
30.11.2016, in rounds 4 and 5 until 31.12.2016 and in round 6 until 1.31.2017. The choice to
omit older data was made mainly based on our empirical experience of modelling
electricity load - too little data would not be enough to create a stable model and too much data
would blur the model with outdated information (the dynamics of electricity load
consumption is likely to change over years).

\paragraph{Trend Variable} The only difference between the 2 mentioned models is that one
uses a trend variable in addition to the original ones. This is defined as follows: a
value of 1 for the year 2003, value of 2 for 2004, ... and value of 15 for 2017. The
usefulness of such a variable was, in some of the target variables, obvious to the naked
eye.

\begin{figure}[!ht]
\label{Fig_TIM_Schema}
\caption{Schematic diagram of TIM.}
\centering
\includegraphics[width=11cm, height=4cm]{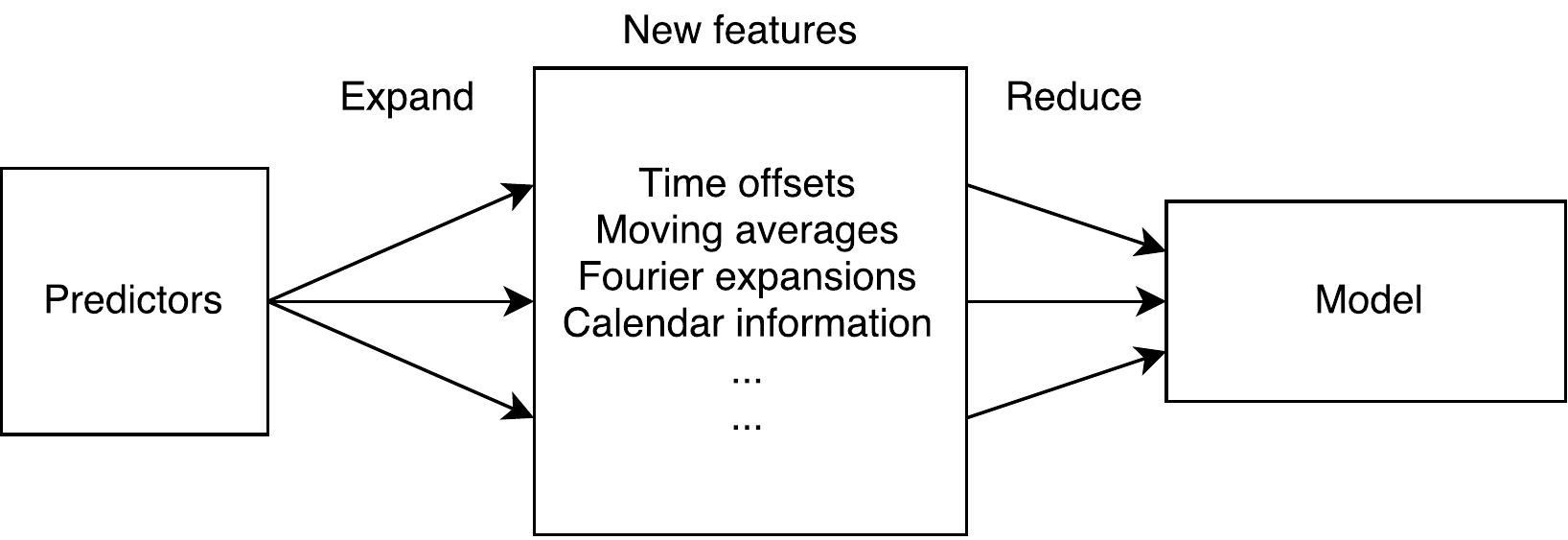}
\end{figure}

\paragraph{Automatic Model Building} The model itself was built using a technique we call
Tangent Information Modeller (TIM). This is a technique where a huge amount of different
transformations is created from the original variables to capture different nonlinear
dependencies and then their number is reduced via an efficient subset selection algorithm.
Transformations such as day of the week, month, moving averages, Fourier expansions,
interactions etc. are all created by searching through the big parameter spaces and
selecting only the most significant from them. Fig.~\ref{Fig_TIM_Schema} depicts this
schema. The resulting model is then a model linear in parameters but nonlinear in its
expressions. Each model building effort therefore optimizes both, the model structure and
its corresponding parameters. Different measures to avoid over-fitting are also taken,
mainly recent results from Information Criteria and Bayesian Statistics
\cite{Konishi2008}. This whole process is fast and automatic, requiring no fine-tuning.
Some other efforts concerning automation in time series modelling are outlined in
\cite{Ziel2016, Hyndman2008, Hyndman2002}.

\noindent For all time series, TIM generated an individual model for each hour of a day.
For the sake of simplicity we refer to this set of models as a model.

\begin{figure}[!ht]
\label{Fig_Model_Deployment}
\caption{General strategy for GEFCom 2017.}
\centering
\includegraphics[width=11cm, height=5cm]{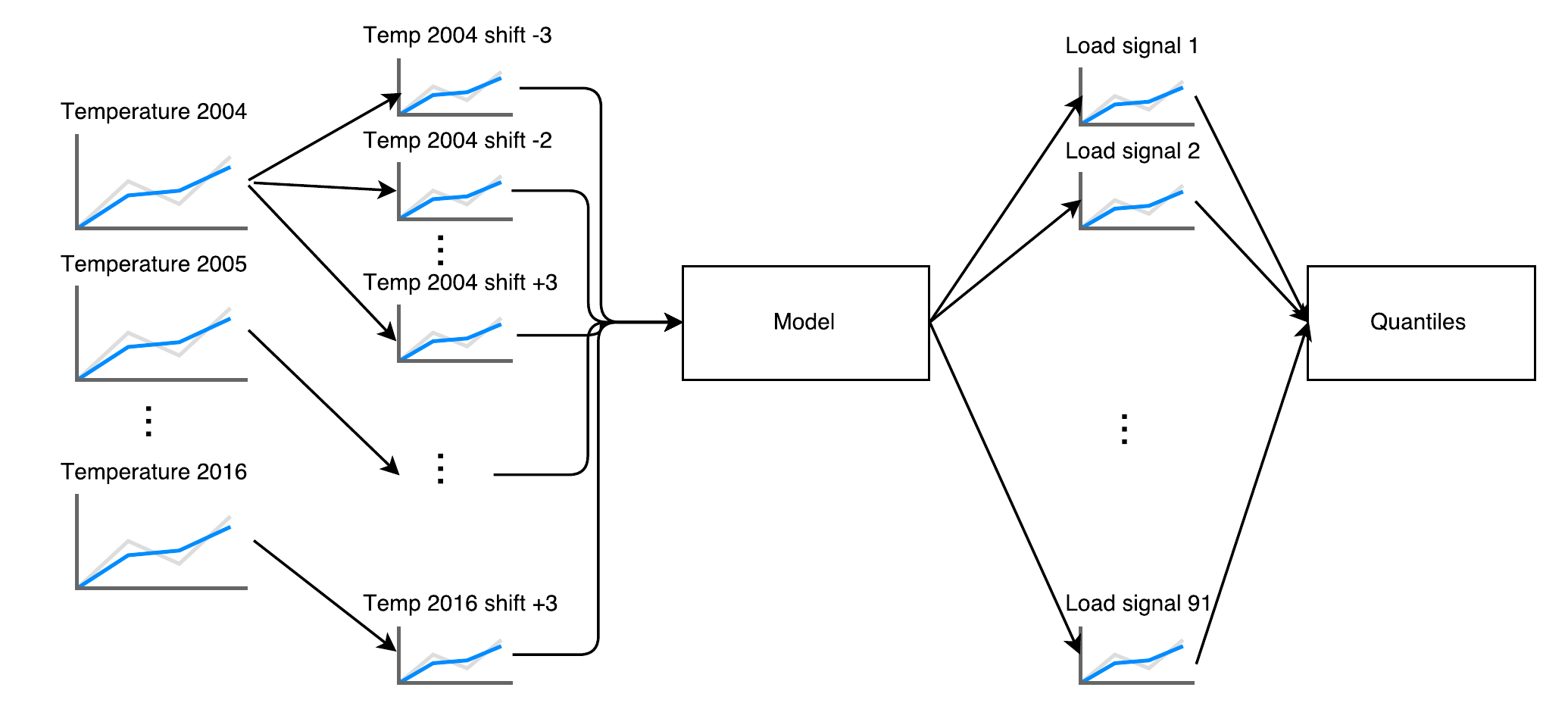}
\end{figure}

\noindent Two sets of quantiles were generated for each model - 'trend' and 'non-trend'
ones. In the 1st, 2nd and 6th round we sent only quantiles from the model containing a
trend and in the 3rd, 4th and 5th round we averaged the quantiles of both models resulting
in an ensemble approach. This decision was taken because after getting partial results
from the first rounds we thought we could improve our performance by weakening the 'trend
effect' (and vice versa in the last round). The whole strategy is visualized in the
Fig.~\ref{Fig_Model_Deployment}.
% TODO: perhaps add somewhere here that TIM is designed for day-ahead forecasts

\section{Results}

\noindent The pinball loss function was used to evaluate the models. Let $\tau$ be the
target quantile, $y$ the real value and $z$ the quantile forecast, then $L_\tau$, the
pinball loss function, can be written as:
\begin{equation}
\begin{array}{l@{}l} L_\tau(y,z) &{}= (y-z)\tau,  \   y\ge z \\ &{}=(z-y)(1-\tau), \   z>y. \end{array}
\end{equation}
For each zone in each round, the pinball loss function was calculated. These were then
compared to the vanilla bench-mark model and relative improvement over the model was
calculated and used as score. The final score for each round is the relative improvement
over the vanilla model averaged across the all zones.

\begin{table}[ht]
\centering
\caption{Scores of different approaches in individual rounds and their corresponding rankings.
Numbers in bold indicate submitted results.}
\label{Tab_Rankings}
\begin{tabular}{|l|c|c|c|c|c|c|c|c|}
	\hline
	Approach         & R1      & R2      & R3      & R4     & R5     & R6       & Mean & Rank \\
	\hline
	Submitted       & 14.55 & 18.92 & 9.30  & 9.86 & 9.74 & 14.92  & 12.88 & 1    \\
	With trend    & \bf{14.55} & \bf{18.92} & 18.92 & 3.28 & 3.16 & \bf{14.92}  & 12.29 & 1    \\
	No trend & -7.20 & -7.08 & -7.08 & 9.96 & 9.84 & -17.14 & -3.12 & 11   \\
	Ensemble      & 6.94  & 9.30  & \bf{9.30}  & \bf{9.86} & \bf{9.74} & 3.47   & 8.17  & 5   \\
	\hline
\end{tabular}
\end{table}

\noindent Table 1 compares 3 different strategies, the trend model, the non-trend model
and their ensemble. The results submitted to the competition are in bold font. The model
with the trend variable was used in rounds 1, 2 and 6. The ensemble of the trend and the
non-trend models was then used in rounds 3, 4 and 5. The results show that having the
trend variable was crucial for winning first place in the competition. However, in
March (rounds 4 and 5) we would have been better off without it.

\section{Discussion}
\paragraph{Automation and Trend Variable}
It was interesting to observe that using models with the trend variable in all rounds
would have also resulted in winning the competition. This can be attributed to the fact
that TIM has the ability to automatically exclude or significantly suppress variables that may
worsen the prediction quality. We could therefore add the trend variable to all of the 10
scenarios and let TIM include or exclude the variable. A closer inspection of the
generated models revealed that in one of the zones the trend variable was completely
excluded while it varied in importance in some others. The whole competition could have been
won in an automatic fashion because both, the underlying models and the quantiles were
estimated without any tuning.

\noindent  Table~\ref{Tab_ComparisonWithVanilla} gives more detailed results concerning
the performance of TIM with the trend variable. These results show that our strategy
consistently outperforms the Vanilla model with very few exceptions. We credit this to
usage of individual underlying model structures for each of the time series. The Vanilla
model, contrastingly, is a fixed structure where only its parameters are estimated from
data. The results suggest that a data-driven individual model structure is perhaps a key to
successful automated modelling.

\paragraph{Responsiveness to New Data} It is worth mentioning that our quantile forecasts
would differ only slightly with the new training data available. This is due to the fact
that shuffled temperatures plugged into the model will not change as they are all
historical values. It was therefore perfectly possible to send quantile forecasts for the
entire year 2017 right at the beginning of the competition. The only thing that may change is
an underlying model, but the strategy should be robust enough that compared to the 3 years
used for training, some small amount of additional training data should not make a
substantial difference.

\noindent Moreover, consider a situation where the temperature in half of January gets
unusually high resulting in a lowered electricity consumption. This would suggest that the
temperature in the other half of January might be high too and therefore we should lower
our quantiles for consumption as well. However, this represents only a small portion of
training data in our underlying model and does not affect temperatures plugged into it at
all, resulting in our model not being able to capture the situation very well. The
question is how to adjust the strategy so models can react to new data faster without
loosing generality.

\paragraph{Usage of Hierarchical Information} Another space for an improvement could be
the “Massachusetts-total" dataset and “Total" dataset. These electricity loads are in
reality a sum of 3 zonal Massachusetts loads and the sum of all loads, respectively. We
completely ignored this connection and built a separate model for each. The only thing
that takes this into account is usage of more temperatures for the Massachusetts-total.
There certainly exists a way to improve results using some sort of hierarchical modelling,
for example as in \cite{Hydman2011}.

\paragraph{Quantile Estimation Techniques} Estimating quantiles directly using, e.g.,
quantile regression as in \cite{Gaillard2016} is also an interesting topic for further
work.

\section{Conclusion}
\noindent This study shows that designing a highly competitive
automatic model building strategy is possible. Full automation brings some interesting
benefits. It is worth emphasizing that a business user with a limited mathematical
background could use TIM in an automatic fashion and obtain the same results. A robust
modelling strategy with zero degrees of freedom is also key to forecasting at scale in
machine-to-machine scenarios where hundreds or thousands of different time series need to
be predicted. TIM is a crucial building block to such a large-scale forecasting system.

\noindent Detecting long-term trends in the historical data is possible and could be
considered as future work. An additional rule for a long-term trend detection is devised
and added to TIM.

\noindent In addition, the applications of TIM are not limited to electricity load
forecasting. At the time of writing, TIM has been extensively tested on several hundreds of
data sets from diverse domains of the energy industry (electricity load, gas consumption, wind
and solar production, district cooling, etc.).

\begin{table}[ht]
\centering
\caption{Performance of TIM with the trend variable over different regions of New England
and rounds. Abbreviations stand for Connecticut, Maine, New Hampshire, Rhode Island,
Vermont, WC Massachusetts, SE Massachusetts, NE Massachusetts and Boston, Total of
Massachusetts and Total of all respectively.}
\label{Tab_ComparisonWithVanilla}
\begin{tabular}{|l|lllllllll|}
\hline
      & R 1     &         &          & R 2     &         &          & R 3     &         &         \\
\hline
      & TIM    & Bench & Score    & TIM    & Bench & Score    & TIM    & Bench & Score   \\
CT    & 99.46  & 114.88  & 13.42  & 86.59  & 115.72  & 25.17  & 86.59  & 115.72  & 25.17 \\
ME    & 24.83  & 36.95   & 32.79  & 22.91  & 29.11   & 21.30  & 22.91  & 29.11   & 21.30 \\
NH    & 38.86  & 41.91   & 7.29   & 34.32  & 35.34   & 2.88   & 34.32  & 35.34   & 2.88  \\
RI    & 19.86  & 23.32   & 14.85  & 17.03  & 24.18   & 29.56  & 17.03  & 24.18   & 29.56 \\
VT    & 19.34  & 22.44   & 13.80  & 14.9   & 15.49   & 3.81   & 14.9   & 15.49   & 3.81  \\
WCMASS   & 44.44  & 50.58   & 12.14  & 46.34  & 60.32   & 23.18  & 46.34  & 60.32   & 23.18 \\
SEMASS   & 40.71  & 44.11   & 7.72   & 40.95  & 50.69   & 19.22  & 40.95  & 50.69   & 19.22 \\
NEMASSBOST  & 66.22  & 77.85   & 14.94  & 62.95  & 81.02   & 22.30  & 62.95  & 81.02   & 22.30 \\
MASS  & 148.36 & 170.2   & 12.83  & 149.88 & 190.36  & 21.26  & 149.88 & 190.36  & 21.26 \\
TOTAL & 339.43 & 402.68  & 15.71  & 313.02 & 401.51  & 22.04  & 313.02 & 401.51  & 22.04 \\
\hline
      & R 4     &         &          & R 5     &         &          & R 6     &         &         \\
\hline
      & TIM    & Bench & Score    & TIM    & Bench & Score    & TIM    & Bench & Score   \\
CT    & 100.48 & 98.91   & -1.59  & 100.48 & 98.8    & -1.70  & 53.69  & 55.11   & 2.58  \\
ME    & 26.49  & 23.96   & -10.58 & 26.49  & 23.88   & -10.95 & 16     & 29.71   & 46.15 \\
NH    & 27.54  & 29.43   & 6.42   & 27.54  & 29.64   & 7.08   & 17.75  & 16.74   & -6.01 \\
RI    & 21.31  & 21.54   & 1.08   & 21.31  & 21.53   & 1.04   & 10.7   & 11.19   & 4.39  \\
VT    & 16.19  & 21.07   & 23.17  & 16.19  & 20.92   & 22.62  & 11.84  & 17.23   & 31.27 \\
WCMASS   & 54.92  & 55.43   & 0.92   & 54.92  & 55.25   & 0.60   & 30.86  & 34.91   & 11.59 \\
SEMASS   & 46.32  & 49.62   & 6.66   & 46.32  & 49.51   & 6.45   & 28.61  & 34.19   & 16.32 \\
NEMASSBOST  & 72.18  & 73.32   & 1.56   & 72.18  & 73.16   & 1.35   & 38.37  & 44.41   & 13.60 \\
MASS  & 170.68 & 175.86  & 2.94   & 170.68 & 175.86  & 2.94   & 86.75  & 106.5   & 18.55 \\
TOTAL & 344.05 & 351.89  & 2.23   & 344.05 & 351.7   & 2.17   & 180.98 & 202.83  & 10.77 \\
\hline
\end{tabular}
\end{table}

\section*{References}

\bibliography{mybibfile}

\end{document}